\documentclass[runningheads]{llncs}

\usepackage{eccv}

\usepackage{eccvabbrv}
\usepackage{enumitem}
\usepackage{graphicx}
\usepackage{booktabs}
\usepackage{comment}
\usepackage[accsupp]{axessibility}  
\usepackage{bm}
\usepackage{algorithm}
\usepackage{algpseudocode}
\usepackage{multirow}
\usepackage{makecell}
\usepackage{hyperref}
\usepackage{orcidlink}

\usepackage{caption}
\DeclareCaptionFormat{myformat}{\fontsize{6}{6}\selectfont#1#2#3}
\newcommand\ExamplesCaption[1]{%
  \captionsetup{format=myformat}%
  \caption{#1}}
\newcolumntype{C}[1]{>{\centering\let\newline\\\arraybackslash\hspace{0pt}}m{#1}}

\begin{document}

\title{U-Sketch: An Efficient Approach for Sketch to Image Diffusion Models} 

\titlerunning{U-Sketch}

\author{Ilias Mitsouras
\and
Eleftherios Tsonis
\and
Paraskevi Tzouveli
\and
Athanasios Voulodimos}

\authorrunning{Mitsouras et al.}

\institute{Artificial Intelligence and Learning Systems Laboratory\\School of Electrical and Computer Engineering\\National Technical University of Athens\\
\email{\{iliasmits, etsonis\}@ails.ece.ntua.gr}}

\maketitle

\begin{abstract}

  Diffusion models have demonstrated remarkable performance in text-to-image synthesis, producing realistic and high resolution images that faithfully adhere to the corresponding text-prompts. Despite their great success, they still fall behind in sketch-to-image synthesis tasks, where in addition to text-prompts, the spatial layout of the generated images has to closely follow the outlines of certain reference sketches. Employing an MLP latent edge predictor to guide the spatial layout of the synthesized image by predicting edge maps at each denoising step has been recently proposed. Despite yielding promising results, the pixel-wise operation of the MLP does not take into account the spatial layout as a whole, and demands numerous denoising iterations to produce satisfactory images, leading to time inefficiency.
  To this end, we introduce \mbox{U-Sketch}, a framework featuring a U-Net type latent edge predictor, which is capable of efficiently capturing both local and global features, as well as spatial correlations between pixels. Moreover, we propose the addition of a sketch simplification network that offers the user the choice of preprocessing and simplifying input sketches for enhanced outputs. The experimental results, corroborated by user feedback, demonstrate that our proposed U-Net latent edge predictor leads to more realistic results, that are better aligned with the spatial outlines of the reference sketches, while drastically reducing the number of required denoising steps and, consequently, the overall execution time.
  
\end{abstract}

\vspace*{-2.2\baselineskip}
\section{Introduction}
\label{sec:intro}
Freehand sketches serve as means of expressing the creativity and imagination of individuals, providing an abstract and relatively simple way of capturing and depicting different aspects of the highly detailed modern world. In this context, the task of transforming them into realistic images, is gaining increasing importance and offers individuals a way to see their imagination come alive. Despite its significance, the task of sketch-to-image synthesis poses a great challenge, as it tries to bridge the gap between the abstract nature of sketches and the rich details of real-world images. This connection requires methods capable of understanding and extrapolating features from the sparse information provided by sketches.

\begin{figure}[t]
    \centering
    \vspace*{-0.3\baselineskip}
    \includegraphics[width = \textwidth]{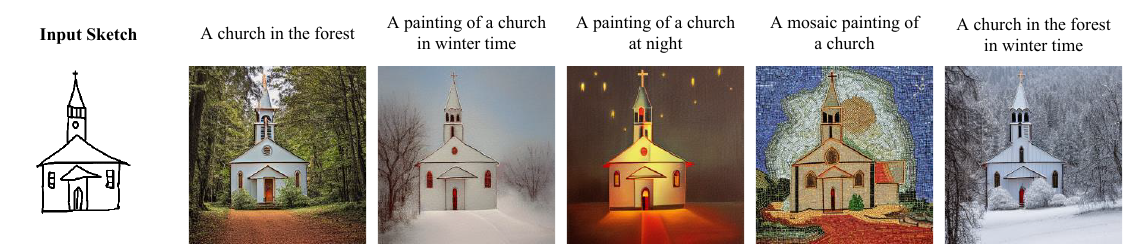}
    \vspace*{-2.7\baselineskip}
\end{figure}

In this context, various GAN-based methods have been proposed to tackle the task of sketch-to-image synthesis \cite{9157030, ghosh2019interactive, liu2020unsupervised, wang2021sketch, zhu2020unpaired, SktechMate, SketchyGan}. Despite their promising results, these methods are limited by the absence of textual guidance, which leads to lack of versatility in the synthesized images. These limitations, in combination with the large amount of paired data that most of them require for training, necessitate the exploration of different methods to tackle the task.

In recent years, diffusion models \cite{dhariwal2021diffusion, ho2020denoising, CascadedDiffusionModels, song2020generative} have garnered attention for their ability to generate high-fidelity images based on corresponding textual descriptions, exhibiting a clear superiority over GANs in text-to-image synthesis tasks \cite{rombach2022highresolution}. Consequently, it is only normal to examine their performance in the task of sketch-to-image synthesis. With this notion in mind, we propose, to the best of our knowledge, the first framework for sketch-guided text-to-image synthesis utilizing a U-Net architecture \cite{ronneberger2015unet} in conjunction with diffusion models.

Our method comprises a U-Net latent edge predictor which extracts and encapsulates spatial pixel correlations, in contrast to recent work of Voynov \etal \cite{voynov}, which utilizes a per-pixel MLP architecture. The U-Net latent edge predictor takes as input a vector of concatenated activations, extracted from the intermediate layers of the denoising U-Net used in Stable Diffusion \cite{rombach2022highresolution} at a certain denoising step and outputs a prediction of the edges of the synthesized image. This prediction is then used to guide the synthesis process, so that the generated image faithfully adheres to the outline of the input sketch. The whole guidance procedure takes place during inference time and thus doesn't require any further training of the pre-trained diffusion model used as backbone. 

In addition, our framework incorporates a sketch simplification network \cite{10.1145/2897824.2925972, simoserra2017mastering}, which refines and smooths out the edges of the input sketches. In this way, we offer the user the option to preprocess the input before feeding it to the diffusion model.  

The experimental results, as well as feedback received by users, show that our proposed U-Sketch:
\begin{itemize}[topsep = 0pt]
    \item[i)] Produces more realistic and high-fidelity images, that faithfully follow the spatial outline of the input sketches. 
    \item[ii)] Reduces the total number of inference steps required to obtain satisfactory results by~$\sim$80\%. This improvement leads to an equally significant reduction in total execution time.
    \item[iii)] Produces images that are preferred by the vast majority of the public opinion, in terms of realism, edge fidelity and overall structure.
\end{itemize}

\section{Related Work}
Sketch-guided image synthesis has been a topic of great interest in the fields of computer vision and image processing. Early methods primarily relied on traditional image processing techniques (e.g. BoW, descriptors, edge detection) to retrieve images from reference sketches \cite{5674030, inproceedings, 5995460, 6116513, ReEnact, InteractiveSketchDrivenSynthesis, Wang2015Sketchbased3S}. These methods, while showing promising results, suffer from several shortcomings, such as limited expressiveness, difficulty in handling complex sketches, and lack of photorealism.

With the advent of deep learning, researchers started to explore GAN-based approaches to tackle the task of sketch-guided image synthesis. By introducing conditional GANs, Mirza \etal \cite{Mirza} paved the way for many image-to-image translation tasks. To this end, Isola \etal \cite{isola2018imagetoimage} implemented 'pix2pix', a conditional GAN capable of learning a mapping from the domain of the input image to a desired output domain in a supervised manner, based on a dataset of paired samples. The first comprehensive effort to guide image generation based on sketches was carried out by Chen \etal \cite{SketchyGan}, with their proposed SketchyGAN. Their framework consists of a generator, with an encoder-decoder architecture and a discriminator. The network is trained in an adversarial manner, where in addition to the adversarial loss, perceptual losses are incorporated to ensure that the generated images are visually similar to the real images. Subsequently, many other GAN-based methods have been proposed \cite{9157030, ghosh2019interactive, liu2020unsupervised, wang2021sketch, zhu2020unpaired, SktechMate} with some of them focusing explicitly on generating human faces from sketch portraits \cite{DeepFaceDrawing, DeepPencilDrawing}. 

Overall, GAN-based methods constituted the first effective way for addressing the task of sketch-guided image synthesis. However, they do not provide the ability for textual guidance. The absence of textual description in combination with mode collapse \cite{ModeCollapse} and instability issues that are present because of the adversarial training of GANs, has prompted the exploration of new methods, based on diffusion models. These likelihood-based models are capable of seamlessly integrating textual guidance, mitigating the shortcomings associated with GANs and providing more robust image synthesis capabilities.

The utilization of diffusion models for sketch-guided text-to-image synthesis has not yet been widely studied, with very few diffusion-based methods being proposed. Voynov \etal \cite{voynov} implemented an MLP latent edge predictor, which estimates an edge map at each step of the denoising process. The edge map is then used to guide the spatial layout of the generated image based on the reference sketch. The edge predictor is trained in a self-supervised, domain agnostic manner. However, their method requires multiple inference steps to achieve satisfactory results, which still fall short compared to the performance of our proposed U-Sketch. Wang \etal \cite{DiffSketching}, train a diffusion model using a hybrid objective function. This objective consists of an identity loss term calculated between the input image and its reconstructed version after the denoising process and a perceptual loss term defined between the input sketch and the one extra\-cted from the reconstructed image. One notable limitation of their approach is the necessity for retraining the model from scratch, which proves to be both costly and time-intensive. In contrast, our approach leverages a pre-trained diffusion backbone directly and the whole sketch-guidance process takes place at inference time.

\newpage
\section{Method}
In this section we describe our sketch-guidance framework, U-Sketch, along with the individual components it comprises. We start with a brief reference to the basic principles of diffusion models and we continue with the analysis of our proposed method. 

\subsection{Background}\label{background_section}
\textbf{Diffusion Models} \ Diffusion models are a class of generative probabilistic models. They work by gradually adding noise to the original data during the forward diffusion process and then attempting to methodically reverse this process. Given a sample $\mathbf{x}_{0}\sim q(\mathbf{x}_{0})$, a variance schedule $\beta_{t}$, the forward diffusion process can be formulated as a Markov chain of fixed length $T$ ($t = 1, \ldots, T$), as follows:
\begin{align}
    q(\mathbf{x}_{1:T}|\mathbf{x}_{0})&  = \prod\limits_{t = 1}^{T}q(\mathbf{x}_{t} | \mathbf{x}_{t-1}),\\
    q(\mathbf{x}_{t} | \mathbf{x}_{t - 1}) & = \mathcal{N}(\mathbf{x}_{t}; \bm{\mu}_{t} = \sqrt{1 - \beta_{t}}\mathbf{x}_{t - 1}, \mathbf{\Sigma}_{t} = \beta_{t}\mathbf{I}),
\end{align}
Then, during the reverse diffusion process, which is also modelled as a Markov chain, we start from a normal distribution $p(\mathbf{x}_{T}) = \mathcal{N}(\mathbf{x}_{T}; \mathbf{0}, \mathbf{I})$ and try to estimate the transition kernels $p_{\theta}(\mathbf{x}_{t - 1}| \mathbf{x}_{t})$. This can be done by training a denoising autoencoder ${\boldsymbol{\epsilon}}_{\theta}(\mathbf{x}_{t}, t)$  which estimates a denoised variant $\mathbf{x}_{t - 1}$ of its input latent $\mathbf{x}_{t}$. Finally using a sequence of these denoising autoencoders for $t = T, \ldots, 1$ we can sample from the desired distribution $\mathbf{x}_{0}\sim p_{\theta}(\mathbf{x}_{0})$.

Many methods have been proposed for the implementation and acceleration of reverse diffusion process. Song Y. \etal \cite{song2021scorebased} model the forward process as a SDE and then try to sample by solving the reverse-time SDE, while Song J. \etal \cite{song2022denoising} use a non-Markovian formulation of the forward and reverse processes, which allows sampling using only a subset of the total steps $T$.

\subsection{U-Net Latent Edge Predictor}
At the core of the proposed sketch-guidance framework is a \textbf{U-Net} latent edge predictor $\mathbf{U}_{\mathrm{LEP}}$, that estimates the edge map of a latent representation $\mathbf{z}_{t}$ of an image at a given step of the denoising process. Following the example of \cite{Barachuk}, the network takes as input a vector $\mathbf{F}$ of concatenated activations $\boldsymbol{l}_{i}, i = 1\ldots, n$, extracted from the intermediate levels of the denoising U-Net ${\boldsymbol{\epsilon}}_{\theta}$ used in Stable Diffusion, along with the noise level $\mathbf{t}$ and its positional encoding $\mathbf{p} = [\sin(2\pi\mathbf{t}\cdot 2^{-i})], i = 0, \ldots, p$, as shown below:
\begin{equation}
    \mathbf{F}(\mathbf{z}_{t}|t, \mathbf{y}) = \left[\boldsymbol{l}_1(\mathbf{z}_{t}  |  t, \mathbf{y}), \ldots, \boldsymbol{l}_n(\mathbf{z}_{t}  |  t, \mathbf{y}),\mathbf{t},\mathbf{p} \right].\label{f_vector}
\end{equation}
Since the activation maps $\boldsymbol{l}_{i}$ come from different layers of the denoising U-Net, they may have varying dimensions. To overcome this issue, before their concatenation we resize them, along channel axis, to match the dimensions of the original input. 

Based on vector $\mathbf{F}$, the edge predictor outputs an estimation $\hat{\mathbf{{E}}}(\mathbf{z}_{t}|t, \mathbf{y}) = \mathbf{U}_{\mathrm{LEP}}( \mathbf{F}(\mathbf{z}_{t}|t, \mathbf{y}))$ of the edge map of the latent representation $\mathbf{z}_{t}$ of the synthesized image, at the current denoising step $t$. This estimation can be later used for the guidance of the synthesis process, based on its similarity with the target edge map.

The use of a U-Net type architecture in our latent edge predictor aims to take advantage of its convolutional nature, which enables it to capture and extract spatial correlations between the pixels of the input tensors. Taking into account that our goal is to estimate spatial outlines, in combination with the fact that edges are in general observed in areas with intense discontinuities in pixel values, it is reasonable to utilize architectures that are capable of processing the input tensors as whole and not in a per-pixel way.  

The training procedure of the U-Net latent edge predictor is similar to the one used in \cite{voynov}. Our training dataset $\boldsymbol{{D}}$ is composed of triplets of the form $(\mathbf{x}, \mathbf{e}, \mathbf{y})$, where $\mathbf{x}$ refers to an input image, $\mathbf{e}$ to its edge map and $\mathbf{y}$ to the corresponding text-prompt. Given triplets of this form, the training process works by adding noise to the latent representation of the input image $\boldsymbol{\mathcal{{E}}}(\mathbf{x})$ and then passing the noisy latent $\mathbf{z}_{t}$ through the denoising U-Net ${\boldsymbol{\epsilon}}_{\theta}$. The extracted intermediate activations are then used to construct the vector $\mathbf{F}$ (eq. \ref{f_vector}), which is  fed to the U-Net latent edge predictor to estimate the edge map $ \hat{\mathbf{{E}}}(\mathbf{z}_{t}|t, \mathbf{y}) $. Finally, the gradient step is taken on the objective function $\mathcal{L}(\hat{\mathbf{{E}}}(\mathbf{z}_{t}|t, \mathbf{y}) , \boldsymbol{\mathcal{E}(\mathbf{e})})$. The whole procedure is presented in detail in Algorithm \ref{training_process}.

\vspace*{-0.5\baselineskip}
\begin{algorithm}[h!]
\caption{U-Net Latent Edge Predictor Training}
    \begin{algorithmic}[1]
    \Require{Datatset $\boldsymbol{{D}}$, noise\_scheduler}\Comment{Require dataset and noise scheduler}
        \Repeat
        \State $(\mathbf{x}, \mathbf{e}, \mathbf{y})\sim \boldsymbol{{D}}$
        \State {Pass $\mathbf{x}, \mathbf{e}$ through the variational encoder to get $\boldsymbol{\mathcal{{E}}}(\mathbf{x})$ and $\boldsymbol{\mathcal{{E}}}(\mathbf{e})$}
        \State $t \sim \mathcal{U}[1, T]$ \Comment{Uniform sampling of noise level}
        \State Add noise to get noisy latent $\mathbf{z}_{t} = \text{noise\_scheduler.add\_noise}(\boldsymbol{\mathcal{{E}}}(\mathbf{x}), t)$
        \State{Pass $\mathbf{z}_{t}$ through the denoising U-Net ${\boldsymbol{\epsilon}}_{\theta}(\mathbf{z}_{t}|t, \mathbf{y})$}
        \State Extract intermediate activations to construct vector $\mathbf{F}(\mathbf{z}_{t}|t, \mathbf{y})$ \Comment{(eq. \ref{f_vector})}
        \State Estimate edge map $\hat{\mathbf{{E}}}(\mathbf{z}_{t}|t, \mathbf{y})  = \mathbf{U}_{\mathrm{LEP}}(\mathbf{F}(\mathbf{z}_{t}|t, \mathbf{y}))$
        \State Take gradient step on
        \vspace*{-0.4\baselineskip}
        \begin{equation}
             \mathcal{L}(\hat{\mathbf{{E}}}(\mathbf{z}_{t}|t, \mathbf{y}) , \boldsymbol{\mathcal{E}(\mathbf{e})}) = \sum\limits_{i, j}\left\| \hat{\mathbf{{E}}}(\mathbf{z}_{t}|t, \mathbf{y}) _{ij} - \boldsymbol{\mathcal{E}}(\mathbf{e})_{ij}\right\|^{2}
            \vspace*{-0.4\baselineskip}
        \end{equation}
    \Until converge
    \end{algorithmic}
    \label{training_process}
\end{algorithm}
\vspace*{-2.2\baselineskip}

\begin{figure}[t]
    \centering
    \includegraphics[width = \textwidth]{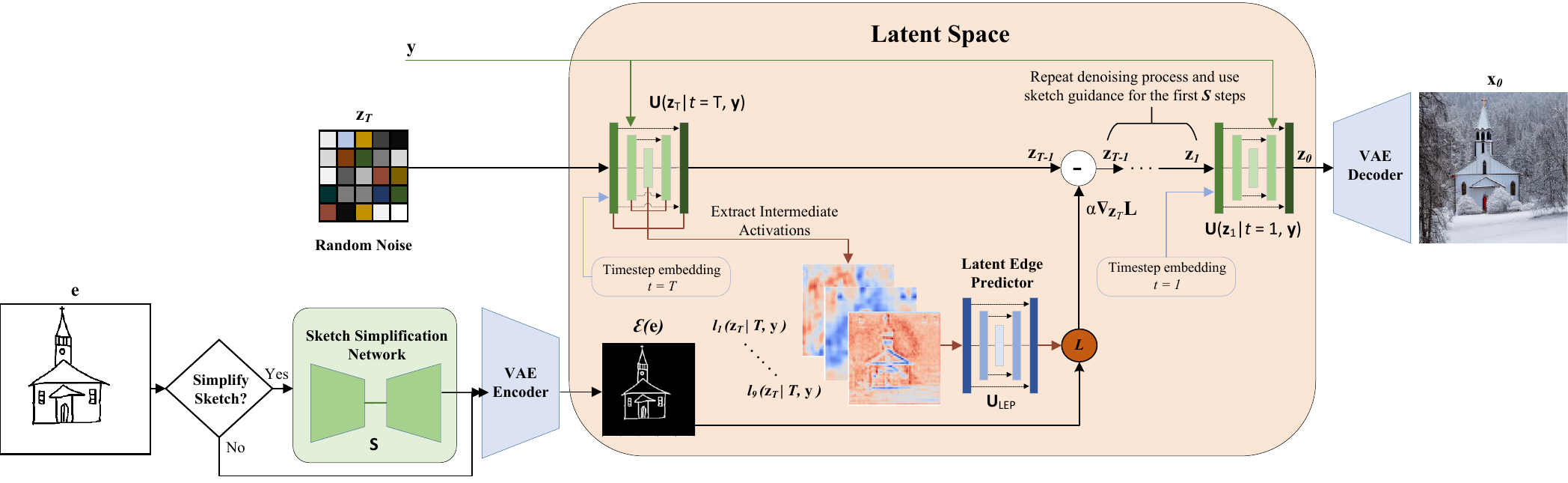}
    \caption{U-Sketch framework. Given an input sketch $\mathbf{e}$ and a text-prompt $\mathbf{y}$, we start by producing random noise $\mathbf{z}_{t}$ and then we iteratively pass it through the denoising autoencoder ${\boldsymbol{\epsilon}}_{\theta}(\mathbf{z}_{t} | t, \mathbf{y})$ to remove noise. During the first $S$ of the total $T$ denoising steps we guide the synthesis process using our U-Net latent edge predictor and sketch's latent representation $\boldsymbol{\mathcal{E}}(\mathbf{e})$, according to the process described in section \ref{synthesis_process_section}. Finally we return  the decoded synthesized image $\mathbf{x}_{0} = \boldsymbol{\mathcal{D}}(\mathbf{z}_{0})$. Sketch simplification network $\mathbf{S}$ offers the user the choice of smoothing the input sketch before feeding it in to the pipeline.}
    \label{synthesis_process_figure}
    \vspace*{-1\baselineskip}
\end{figure}

\subsection{Image Synthesis using U-Net Latent Edge Predictor}\label{synthesis_process_section}
After training our U-Net latent edge predictor, we can now use it to guide the reverse diffusion process. As mentioned in section \ref{background_section}, reverse diffusion consists of a sequence of denoising autoencoders ${\boldsymbol{\epsilon}}_{\theta}(\mathbf{x}_{t}, t)$ for $t = T, \ldots 1$ which estimate a denoised variant of their input $\mathbf{x}_t$. In our implementation, we use Stable Diffusion as the backbone of image synthesis. This model transfers the whole diffusion process in latent space and thus, given a noisy latent representation $\mathbf{z}_{t}$ at step $t$ along with the corresponding text-prompt $\mathbf{y}$, the denoising autoencoder ${\boldsymbol{\epsilon}}_{\theta}(\mathbf{z}_{t} | t, \mathbf{y})$ estimates a denoised version $\mathbf{z}_{t - 1}$ at step $t - 1$.

To guide the aforementioned synthesis process towards the spatial outline of the input sketch $\mathbf{e}$ we use a method similar to \cite{voynov, dhariwal2021diffusion}. More precisely, at each denoising step $t$, given the noisy latent $\mathbf{z}_{t}$ we first estimate its denoised version $\mathbf{z}_{t - 1}$. We, then, use the U-Net latent edge predictor to estimate the corresponding edge map $\hat{\mathbf{{E}}}(\mathbf{z}_{t}|t, \mathbf{y})$, based on the intermediate activations extracted from the denoising U-Net ${\boldsymbol{\epsilon}}_{\theta}$. This estimation is compared to the original edge map and a similarity between the two is defined as follows, 
\begin{equation}
    \mathrm{L}(\hat{\mathbf{{E}}}(\mathbf{z}_{t}|t, \mathbf{y}) , \boldsymbol{\mathcal{E}}(\mathbf{e})) = \|\hat{\mathbf{{E}}}(\mathbf{z}_{t}|t, \mathbf{y}) - \boldsymbol{\mathcal{E}}(\mathbf{e}) \|^{2}. \label{similarity_measure}
\end{equation}
The similarity expresses how closely the spatial layout of the synthesized image at step $t$ adheres to the corresponding layout of the reference sketch. To guide the synthesis process, we calculate the gradient of the similarity measure (eq. \ref{similarity_measure}) and then use it to modify the original estimation $\mathbf{z}_{t -1}$ in the following way,
\begin{equation}
    \mathbf{z}_{t - 1} \leftarrow \mathbf{z}_{t - 1}-\alpha\mathbf{\nabla}_{{\mathbf{z}}_{t}}\mathrm{L}(\hat{\mathbf{{E}}}(\mathbf{z}_{t}|t, \mathbf{y}) , \boldsymbol{\mathcal{E}}(\mathbf{e})),\label{guidance_eq}
\end{equation}
where $\alpha$ is a parameter that controls the sketch-guidance strength and defined as follows,
\begin{equation}
    \alpha = \frac{\|\mathbf{z}_{t} - \mathbf{z}_{t-1} \|_{2}}{\|\mathbf{\nabla}_{\mathbf{z}_{t}}\mathrm{L}(\hat{\mathbf{{E}}}(\mathbf{z}_{t}|t, \mathbf{y}) , \boldsymbol{\mathcal{E}}(\mathbf{e}))\|_{2}}\cdot \beta.\label{guidance_strength_equation}
\end{equation}
According to the above equation, the magnitude of the gradient is relative to the magnitude of the difference between the original estimation and the noisy latent of the previous step. $\beta$ functions as a weighting factor within $\alpha$ to balance the aspects of edge fidelity and realism in the generated images.

The impact of sketch-guidance is present during the first $S$ steps of the reverse diffusion process, while afterwards, and for the rest $T - S$ steps  we lift the sketch's restriction and leave the diffusion process to continue unaffected. 

Finally, our framework incorporates a sketch simplification network $\mathbf{S}$ \cite{simoserra2017mastering} which simplifies and smooths the edges of the sketches. This symmetric convolutional network operates in pixel space and is trained in a hybrid way, utilizing pairs of sketches and their simplified versions along with a discriminator, that aims to distinguish true edges from generated ones. This network offers the choice of preprocessing the input sketch before feeding it to the synthesis pipeline. 

The total sketch-guided synthesis process using our U-Net latent edge predictor is described in Algorithm \ref{synthesis_process_algorithm} and depicted in Figure \ref{synthesis_process_figure}.

\vspace*{-0.7\baselineskip}
\begin{algorithm}[h!]
\caption{Sketch-guided Image Synthesis}
    \begin{algorithmic}[1]
        \Require{($\mathbf{y}, \mathbf{e}$), $S$, $\beta$, noise\_scheduler}
        \State $\mathbf{z}_{T} \sim \mathcal{N}(\mathbf{0}, \mathbf{I})$\Comment{Initialize random noise}
        \State Get latent representation $\boldsymbol{\mathcal{E}}(\mathbf{e})$
        \For{$t = T, \ldots , 1$}
            \State{Calculate $\mathbf{z}_{t - 1}$ based on $\mathbf{z}_{t}$, $\mathbf{z}_{t - 1}= \text{noise\_scheduler.step}(\mathbf{z}_{t}, t, {\boldsymbol{\epsilon}}_{\theta}(\mathbf{z}_{t}|t, \mathbf{y}))$}
            \State{Estimate edge map $\hat{\mathbf{{E}}}(\mathbf{z}_{t}|t, \mathbf{y}) = \mathbf{U}_{\text{LEP}} (\mathbf{F}(\mathbf{z}_{t}|t, \mathbf{y}))$}
            \If{$T - t \leq S$}
            \State{Calculate $\alpha$\Comment{(eq.\ref{guidance_strength_equation})}}
            \State{${{\mathbf{z}}}_{t - 1} \leftarrow \mathbf{z}_{t - 1} - \alpha \mathbf{\nabla}_{\mathbf{z}_{t}}\mathrm{L}(\hat{\mathbf{{E}}}(\mathbf{z}_{t}|t, \mathbf{y}) , \boldsymbol{\mathcal{E}}(\mathbf{e}))$}
            \EndIf
        \EndFor
        \State \textbf{return} $\mathbf{x}_{0} = \boldsymbol{\mathcal{D}}(\mathbf{z}_{0})$\Comment{Pass latent $\mathbf{z}_{0}$ through variational decoder $\boldsymbol{\mathcal{D}}$}
    \end{algorithmic}
    \label{synthesis_process_algorithm}
\end{algorithm}

\vspace*{-1.7\baselineskip}
\section{Experiments and Results}
In this section we delve into the implementation details of our approach and conduct a series of experiments to evaluate the performance of our proposed framework. For the qualitative part of the evaluation we present various results of synthesized images and compare them with the corresponding results obtained by \cite{voynov}. Finally, after taking into account the lack of suitable evaluations metrics, we conduct user studies to quantitatively evaluate our results.
\vspace*{-0.5\baselineskip}
\subsection{Experimental Setup}
\textbf{Model Configuration:} \ To generate high-fidelity images based on text-prompts, we utilize \href{https://huggingface.co/runwayml/stable-diffusion-v1-5/tree/main}{Stable Diffusion v1.5}. The variational encoder used in Stable Diffusion outputs latent representations with a total of 4 channels. 

Our proposed U-Net latent edge predictor adopts the following architecture: the encoder block consists of a sequence of 4 convolutional levels, each containing 2 convolutional layers followed by ReLU activation and a max pooling operation, with 64, 128, 256 and 512 filters, respectively. The bottleneck comprises 2 convolutional layers with 1024 filters each. The decoder block mirrors the encoder's architecture in a symmetric way. The output block is a convolutional layer with 4 output channels. For the MLP latent edge predictor we adopt the architecture \newpage \noindent  used in the original implementation, which consists of 4 fully-connected layers with ReLU activations, followed by batch normalization layers, hidden dimensions 512, 256, 128, 64, and output dimension 4.

The intermediate activations $\boldsymbol{l}_{i}, i = 1, \ldots, n$ are extracted from a total of $n = 9$ different intermediate levels of Stable Diffusion's denoising U-Net, as follows: $\boldsymbol{l}_{1}, \ldots, \boldsymbol{l}_{3} $ are extracted from layers 2, 4 and 8 of the input block, $\boldsymbol{l}_{4}, \ldots, \boldsymbol{l}_{6} $ from layers 0, 1, 2 of the middle block and $\boldsymbol{l}_{7}, \ldots, \boldsymbol{l}_{9} $ from layers 2, 4, 8 of the output block. For the positional encoding we set $p  = 9$.

\noindent \textbf{Training:} \ Our training dataset consists of images collected from ImageNet \cite{ImageNet}. More specifically, we randomly sample 40 images from 150 different classes, to obtain a total of 6000 images. To extract the corresponding edge maps we utilize PiDiNet \cite{su2021pixel}, a convolutional neural network that estimates edges based on pixel difference. The edge maps extracted from PiDiNet are then binarized using a threshold value of 0.5 and the resulting binarized images form the desired edge maps. Finally, to get the textual description of each image we use the name of the class to which the image is included. Both MLP and U-Net latent edge predictors are trained for a total of 10 epochs.

\noindent \textbf{Inference:} \ As discussed in section \ref{synthesis_process_section}, the inference process involves the tuning of several hyperparameters. During our experiments we employ a DDIM scheduler to update the samples at each denoising step, deterministically. For all the experiments involving our proposed U-Net latent edge predictor, the denoising process consists of a total of $T = 50$ steps, while $\beta$ is set to $1.6$ for a total of $S = 0.5\hspace*{1pt}T = 25$ steps. Following thorough experimentation, we deduce that \linebreak values of $\beta$ in range $[1.5, 1.8]$, and values of $S$ in range $[0.45 \hspace*{1pt} T, 0.55 \hspace*{1pt} T]$ lead to satisfactory results, with lower values of $S$ being better suited to sketches of higher spatial complexity. Finally, we set guidance scale equal to 8 for the classifier-free guidance, to increase the adherence to the text-prompt. All the  sketches used during inference are part of the Sketchy Dataset \cite{SketchyDatabase}.

\subsection{Qualitative Evaluation}
\begin{figure}[h!]
\centering
    \begin{subfigure}[h!]{0.49\textwidth}
    \includegraphics[width = \textwidth]{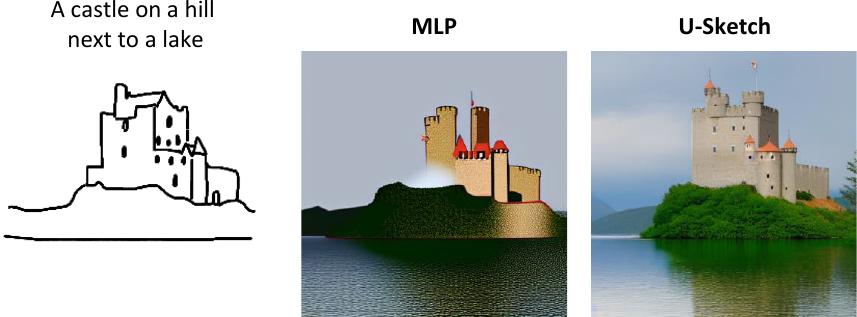}
    \vspace{-0.7\baselineskip}
    \ExamplesCaption{}
    \label{section_4_2_examples_a}
    \end{subfigure}\hspace{0.2\baselineskip}
    \begin{subfigure}[h!]{0.49\textwidth}
    \includegraphics[width = \textwidth]{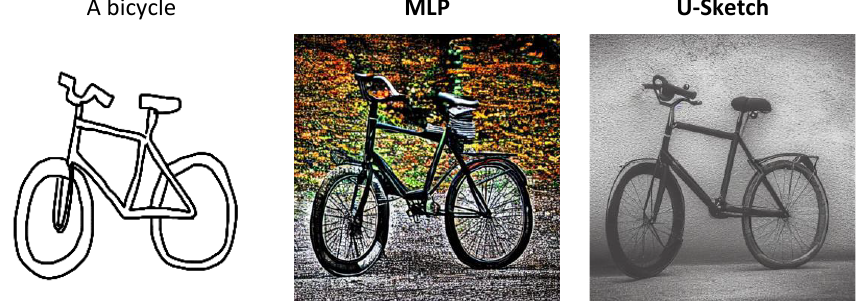}
    \vspace{-0.7\baselineskip}
    \ExamplesCaption{}
    \label{section_4_2_examples_b}
    \end{subfigure}
    \begin{subfigure}[h!]{0.49\textwidth}
    \includegraphics[width = \textwidth]{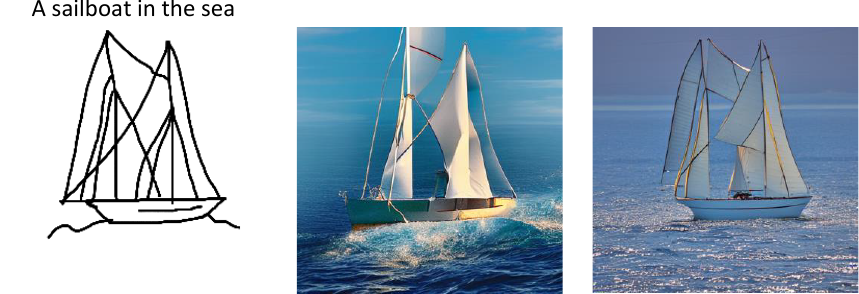}
    \vspace{-0.7\baselineskip}
    \ExamplesCaption{}
    \label{section_4_2_examples_c}
    \end{subfigure}\hspace{0.2\baselineskip}\begin{subfigure}[h!]{0.49\textwidth}
    \includegraphics[width = \textwidth]{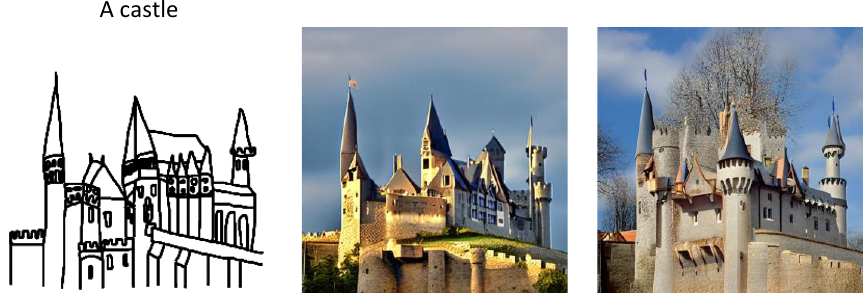}
    \vspace{-0.7\baselineskip}
    \ExamplesCaption{}
    \label{section_4_2_examples_d}
    \end{subfigure}
    \begin{subfigure}[h!]{0.49\textwidth}
    \includegraphics[width = \textwidth]{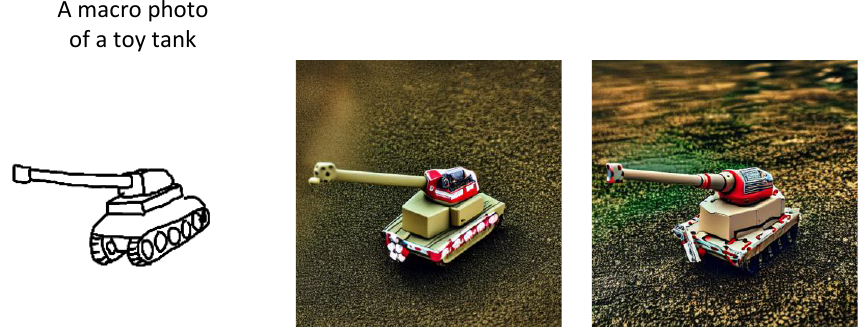}
    \vspace{-0.7\baselineskip}
    \ExamplesCaption{}
    \label{section_4_2_examples_e}
    \end{subfigure}\hspace{0.2\baselineskip}\begin{subfigure}[h!]{0.49\textwidth}
    \includegraphics[width = \textwidth]{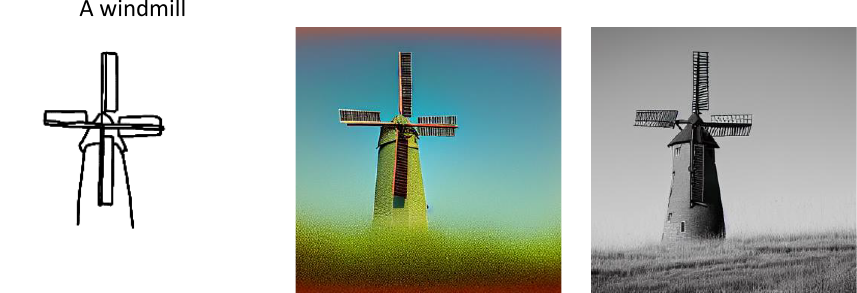}
    \vspace{-0.7\baselineskip}
    \ExamplesCaption{}
    \label{section_4_2_examples_f}
    \end{subfigure}
     \begin{subfigure}[h!]{0.49\textwidth}
    \includegraphics[width = \textwidth]{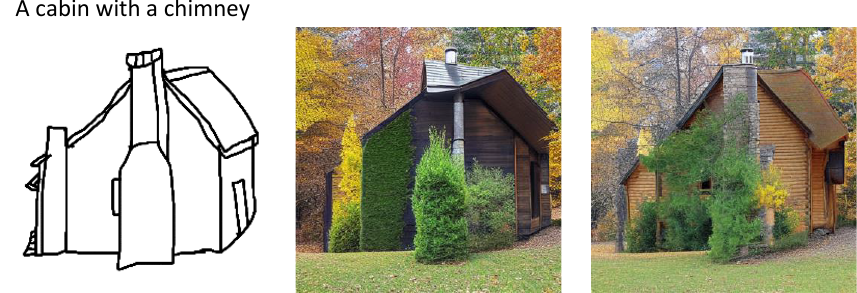}
    \vspace{-0.7\baselineskip}
    \ExamplesCaption{}
    \label{section_4_2_examples_g}
    \end{subfigure}\hspace{0.2\baselineskip}\begin{subfigure}[h!]{0.49\textwidth}
    \includegraphics[width = \textwidth]{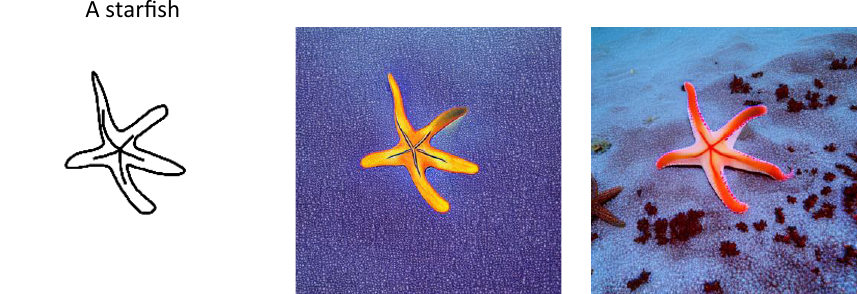}
    \vspace{-0.7\baselineskip}
    \ExamplesCaption{}
    \label{section_4_2_examples_h}
    \end{subfigure}
    \begin{subfigure}[h!]{0.49\textwidth}
    \includegraphics[width = \textwidth]{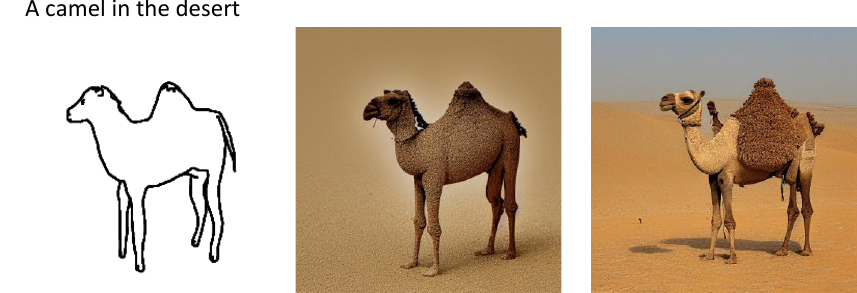}
    \vspace{-0.7\baselineskip}
    \ExamplesCaption{}
    \label{section_4_2_examples_i}
    \end{subfigure}\hspace{0.2\baselineskip}\begin{subfigure}[h!]{0.49\textwidth}
    \includegraphics[width = \textwidth]{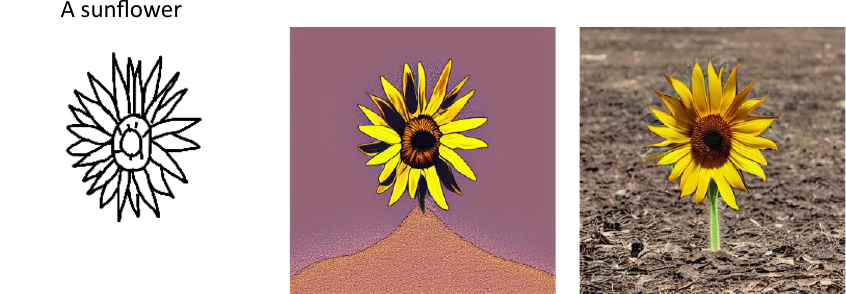}
    \vspace{-0.7\baselineskip}
    \ExamplesCaption{}
    \label{section_4_2_examples_j}
    \end{subfigure}
    \caption{Sketch-guided text-to-image synthesis examples. For each triplet we have from left to right: the reference sketch along with the text-prompt, the image generated using the MLP and the image generated using our proposed U-Sketch framework.}
    \label{comparative_analysis_figs}
    \vspace*{-1.4\baselineskip}
\end{figure}

For our first experiment we conduct a comparative qualitative evaluation between the results obtained by the MLP and our proposed U-Net latent edge predictor. In Figure \ref{comparative_analysis_figs} we present a series of sketch-guided text-to-image synthesis examples. Each triplet of images is structured from left to right: the input sketch with the corresponding text-prompt, the image produced by the MLP and the image produced by our U-Net latent edge predictor, using the same seed number for both. 

Upon scrutinizing the images shown in Figure \ref{comparative_analysis_figs}, we discern that our proposed framework outperforms the MLP, in terms of both realism and edge fidelity. Specifically, our framework generates images that exhibit a higher degree of realism, with the depicted objects resembling more closely their real-world counterparts. Additionally, the background in which these objects are situated, demonstrates wide versatility, blending into real-world sceneries. At the same time, our results are, for the most part, better aligned with the spatial layout of the reference sketches and encapsulate the character of the textual description.

\newpage

\noindent  Even in cases where the MLP manages to capture the edges of the input sketches more effectively (e.g. Figure \ref{section_4_2_examples_h}, \ref{section_4_2_examples_i}), this spatial adherence comes at the expense of realism, with the synthesized images exhibiting a more anime-like appearance, with a solid and relatively plain background. 

This behaviour is attributed to the per-pixel nature of the MLP, which treats each pixel individually, disregarding the correlations between them. In contrast, our U-Net architecture can effectively capture these type of correlations. This spatial knowledge enables the U-Net to provide better estimations of the edge maps at earlier steps of the denoising process. As a result, the sketch-guidance process is performed more effectively, while exerting minimal impact on the back\-ground elements of the synthesized images. Generally, our framework tends

\newpage
\noindent  to prioritize the preservation of realism over attaining absolute spatial precision. Thereby, when the input sketches exhibit a high level of complexity (e.g. Figure \ref{section_4_2_examples_c}, \ref{section_4_2_examples_d}), our framework generates images, that despite being slightly less accurate in terms of the spatial outline, possess a heightened sense of realism.

\begin{figure}[h!]
    \begin{subfigure}{\textwidth}
    \begin{minipage}{0.28\textwidth}
        \ExamplesCaption{}
        \label{section_4_2_examples_a_more_steps}
    \end{minipage}\hspace*{-1.4cm}
    \begin{minipage}{0.67\textwidth}
        \includegraphics[width = \textwidth]{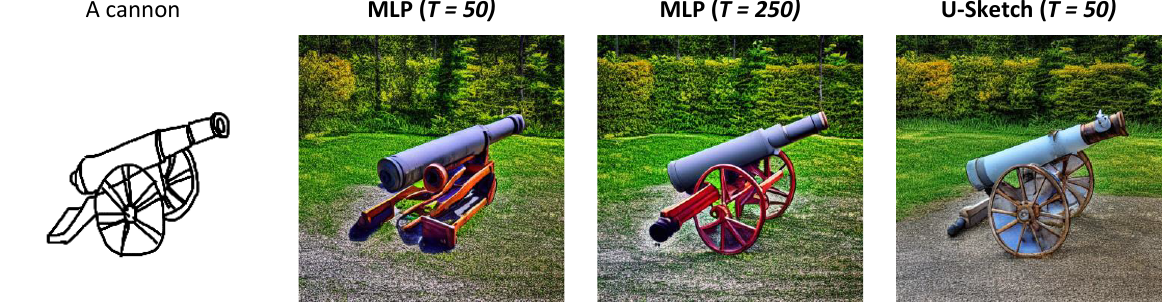}
    \end{minipage}
    \vspace*{0.15\baselineskip}
    \end{subfigure}
    \vspace*{-0.2\baselineskip}
   \begin{subfigure}{\textwidth}
    \begin{minipage}{0.28\textwidth}
        \ExamplesCaption{}
        \label{section_4_2_examples_b_more_steps}
    \end{minipage}\hspace*{-1.4cm}
    \begin{minipage}{0.67\textwidth}
        \includegraphics[width = \textwidth]{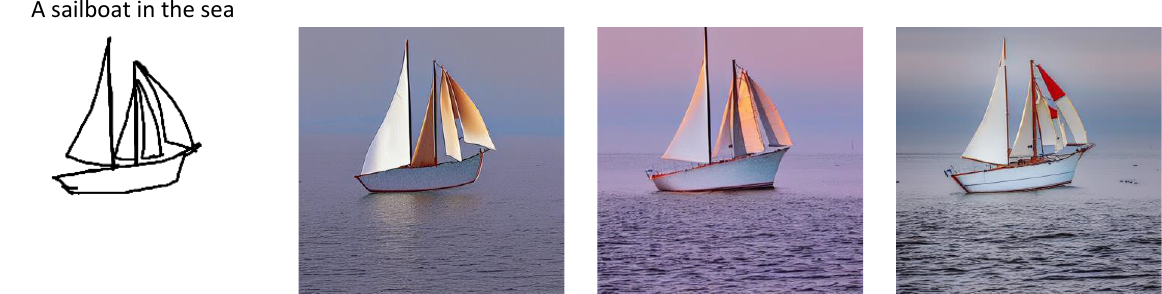}
    \end{minipage}
    \vspace*{0.15\baselineskip}
    \end{subfigure}
    \vspace*{-0.2\baselineskip}
    \begin{subfigure}{\textwidth}
    \begin{minipage}{0.28\textwidth}
        \ExamplesCaption{}
        \label{section_4_2_examples_c_more_steps}
    \end{minipage}\hspace*{-1.4cm}
    \begin{minipage}{0.67\textwidth}
        \includegraphics[width = \textwidth]{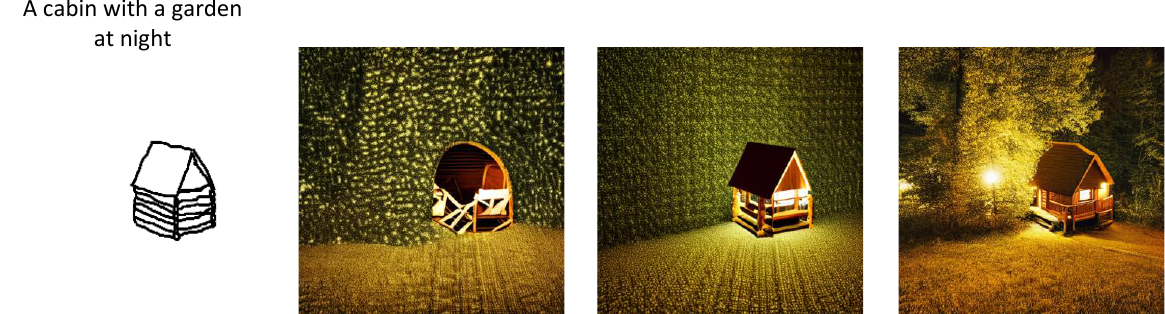}
    \end{minipage}
    \vspace*{0.2\baselineskip}
    \end{subfigure}
    \caption{Sketch-guided text-to-image synthesis examples. For each example we have from left to right: the reference sketch along with the text-prompt, the image generated using the MLP with $T = 50$ and $T = 250$ denoising steps and the image generated using our proposed U-Sketch framework with $T = 50$ denoising steps.}
    \label{comparative_analysis_figs_more_steps}
\end{figure}

At this point, in order to ensure that our comparison is thorough and impartial, we provide additional examples, utilizing more denoising steps for the case of the MLP latent edge predictor. For each example in Figure \ref{comparative_analysis_figs_more_steps} we have from  left to right:  the input sketch with the corresponding text-prompt, the image produced by the MLP using $T = 50$ and $T = 250$ steps and the image produced by our U-Net using $T = 50$ steps. 

Based on these results, we observe that the increase of denoising steps leads to an overall improvement in the performance of the MLP latent edge predictor. The newly synthesized images are better aligned with the input sketches and the borders of their individual components are more efficiently differentiated. Despite this overall improvement, the quality of the images is significantly lower in comparison to those obtained using our U-Net latent edge predictor. It is worth noting that our approach boasts an 80\% reduction in denoising steps and calls to the latent edge predictor. At the same time, as shown in Figure \ref{section_4_2_examples_c_more_steps}, the increase of denoising steps does not guarantee the realism of the final result. Finally, in terms of time efficiency, increasing the number of denoising steps from 50 to 250, results to a proportional increase in the overall execution time. For instance, the total execution time required to produce an image using our framework with 50 denoising steps is approximately 50 seconds on a T4 NVIDIA GPU, whereas when using the MLP latent edge predictor with 250 steps, it increases to approximately 250 seconds.  

\vspace*{-1.1\baselineskip}

\newpage
\subsection{Impact of Sketch Simplification Network}
\begin{figure}[h!]
\centering
    \begin{subfigure}[h!]{0.49\textwidth}
    \includegraphics[width = \textwidth]{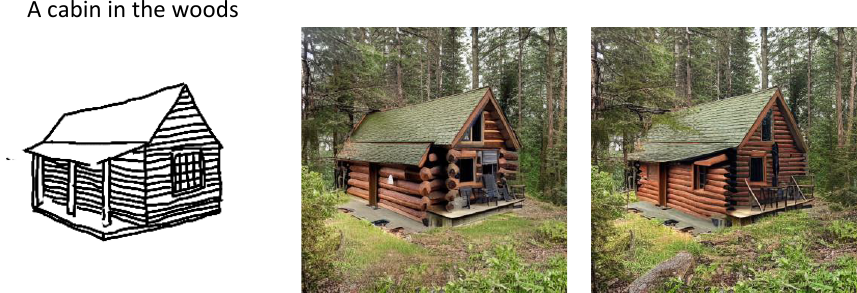}
    \vspace{-0.7\baselineskip}
    \ExamplesCaption{}
    \label{section_4_3_examples_a}
    \end{subfigure}\hspace{0.2\baselineskip}
    \begin{subfigure}[h!]{0.49\textwidth}
    \includegraphics[width = \textwidth]{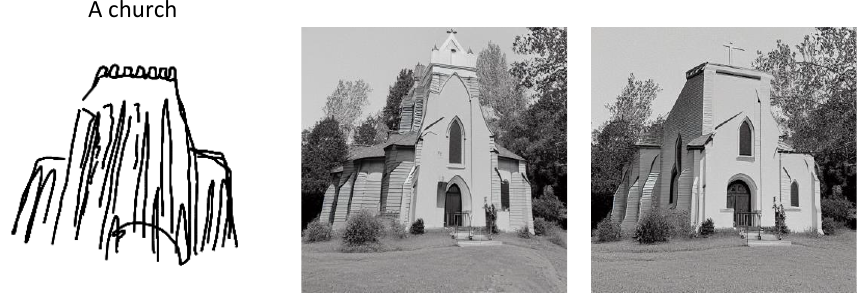}
    \vspace{-0.7\baselineskip}
    \ExamplesCaption{}
    \label{section_4_3_examples_b}
    \end{subfigure}
    \begin{subfigure}[h!]{0.49\textwidth}
    \includegraphics[width = \textwidth]{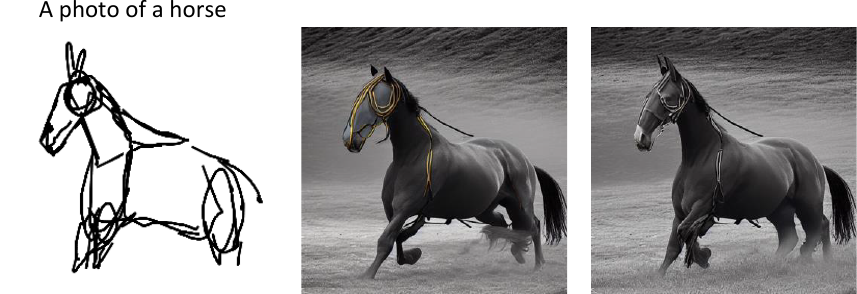}
    \vspace{-0.7\baselineskip}
    \ExamplesCaption{}
    \label{section_4_3_examples_c}
    \end{subfigure}\hspace{0.2\baselineskip}
    \begin{subfigure}[h!]{0.49\textwidth}
    \includegraphics[width = \textwidth]{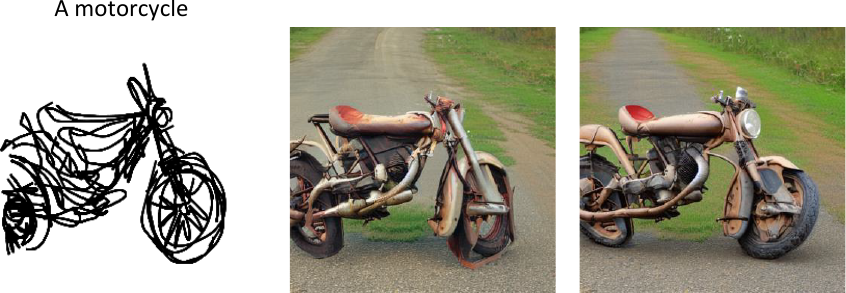}
    \vspace{-0.7\baselineskip}
    \ExamplesCaption{}
    \label{section_4_3_examples_d}
    \end{subfigure}
    \caption{Sketch-guided text-to-image synthesis examples using U-Sketch. For each triplet we have from left to right: the reference sketch along with the text-prompt, the image generated without the use and with the use of the sketch simplification network.}
    \label{sketch_simplification_examples}
    \vspace*{-2.5\baselineskip}
\end{figure}

As mentioned in section \ref{synthesis_process_section}, a fundamental component of our proposed U-Sketch framework is a sketch simplification network, that allows users to simplify and smoothen their sketches. This network is particularly useful in cases of rough and poorly drawn sketches, where it is used to alleviate the impact of the overlapping and indistinct lines to the sketch-guidance process. To assess the contribution of the network to the overall performance of our framework, in Figure \ref{sketch_simplification_examples} we present some results, obtained both with and without its utilization. 

Based on these results, we can conclude that the sketch simplification network enhances the generated images, in various  aspects. From Figure \ref{section_4_3_examples_a} we note that the refinement applied to the input sketch, leads to a better capture of the position and width of its dense parallel lines. Furthermore, in Figure \ref{section_4_3_examples_b} we can clearly observe that the simplification of the input sketch reduces the impact of the intermediate abstract, disorderly lines, resulting in better alignment with the input spatial outlines. Finally, examples \ref{section_4_3_examples_c} and \ref{section_4_3_examples_d} demonstrate an overall improvement after simplifying the input sketches, both in terms of realism and edge fidelity, corroborating the aforementioned observations.

\vspace*{-0.5\baselineskip}

\subsection{Impact of Noise Initialization}
Noise initialization is a pivotal part of the diffusion process, serving as the foundation for generating new images. Despite its random and unstructured nature, noise possesses a basic intrinsic layout that is not obvious with simple observation. By stabilizing the seed and text-prompt of the diffusion process, we generate images where the depicted objects exhibit consistent layout and positioning.

\begin{figure}[t]
    \begin{subfigure}{\textwidth}
    \begin{minipage}{0.12\textwidth}
        \ExamplesCaption{}
        \label{section_4_4_examples_a}
    \end{minipage}\hspace*{-0.32cm}
    \begin{minipage}{0.85\textwidth}
        \includegraphics[width = \textwidth]{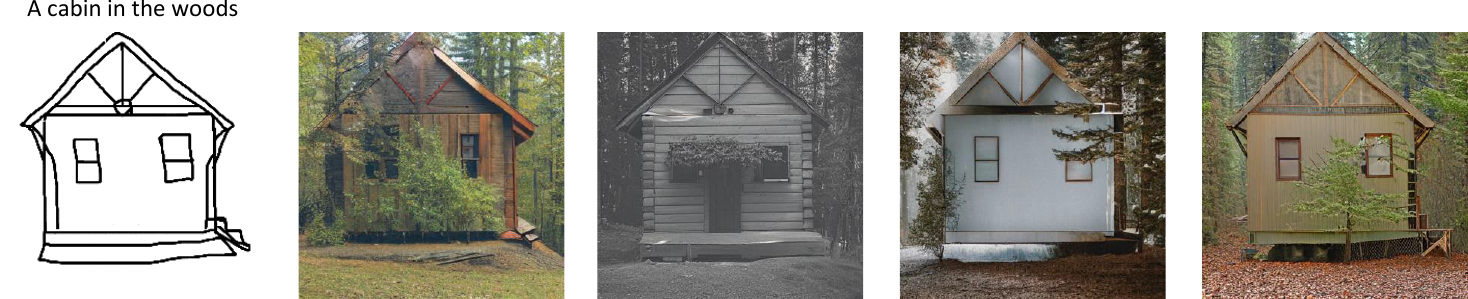}
    \end{minipage}
    \vspace*{0.2\baselineskip}
    \end{subfigure}
    \vspace*{-0.1\baselineskip}
   \begin{subfigure}{\textwidth}
    \begin{minipage}{0.12\textwidth}
        \ExamplesCaption{}
        \label{section_4_4_examples_b}
    \end{minipage}\hspace*{-0.32cm}
    \begin{minipage}{0.85\textwidth}
        \includegraphics[width = \textwidth]{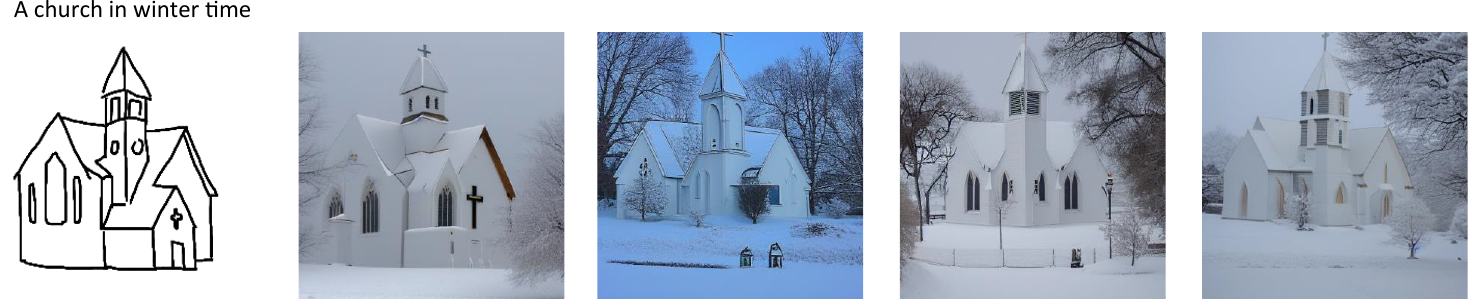}
    \end{minipage}
    \vspace*{0.2\baselineskip}
    \end{subfigure}
    \vspace*{-0.1\baselineskip}
    \begin{subfigure}{\textwidth}
    \begin{minipage}{0.12\textwidth}
        \ExamplesCaption{}
        \label{section_4_4_examples_c}
    \end{minipage}\hspace*{-0.32cm}
    \begin{minipage}{0.85\textwidth}
        \includegraphics[width = \textwidth]{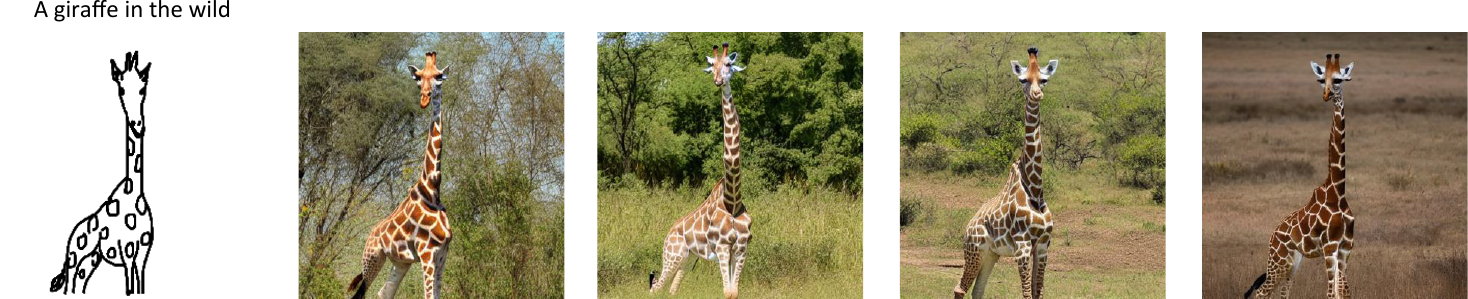}
    \end{minipage}
    \vspace*{0.2\baselineskip}
    \end{subfigure}
    \caption{Sketch-guided text-to-image synthesis examples using U-Sketch and different noise initializations.}
    \label{different_seeds_examples}
    \vspace*{-1.4\baselineskip}
\end{figure}

In this section we aim to examine how the initial random noise affects the sketch-guidance process and the overall results. Figure \ref{different_seeds_examples} presents a series of examples, where for each input sketch and text-prompt, we use 4 different noise initializations to get the resulting images. Based on these results, it is evident that the initial noise significantly affects the generated images, both in terms of background scenery and edge fidelity. More precisely, in examples \ref{section_4_4_examples_a}, \ref{section_4_4_examples_c} the variation of noise initialization leads to images with different background sceneries and color palettes, while maintaining similar spatial layouts for the objects of interest. In contrast, in example \ref{section_4_4_examples_b} apart from the versatility in background, we also observe a divergence in spatial layout, with some initializations leading to better adhering to the outlines of the input sketch (first and fourth generated images), while others to a slight deviation from the reference geometry (second and third generated images).

In conclusion, we can infer that the initial noise used in sketch-guided text-to-image synthesis poses a factor of great importance for the quality and fidelity of the generated images. In cases where its intrinsic layout is better aligned with the spatial outlines of the reference sketch, the guidance process is significantly aided and thus leads to better capturing of this geometry. Conversely, when the initial intrinsic layout diverges substantially from the reference spatial topology, the whole process becomes notably harder, with the resulting images either slightly diverging or completely failing to adhere to the input sketches.

\subsection{Quantitative Evaluation}
Up to this point, our evaluation has primarily focused on qualitative assessments. In this section we proceed to conduct a quantitative evaluation to enrich the performance study of the MLP and our proposed U-Net latent edge predictor.

A major challenge in sketch-guided text-to-image synthesis is the absence of metrics that effectively integrate the aspects of edge fidelity, realism and overall quality of the synthesized images. To overcome this limitation, we conduct two user studies and utilize public opinion to quantify the performance of the two latent edge predictors. Additionally, in order to obtain a rudimentary quantification of the spatial aspect of the synthesized images, we employ recall metric.

\noindent \textbf{Recall:} \ In our study, recall is employed to compare the input sketch with the edge map extracted from the corresponding synthesized image. To extract this edge map we once again utilize PiDiNet, as it has the capability of disregarding some edges that are not of interest (e.g. complex edges that are situated in the background). Considering that our goal is to focus our comparison on the areas containing the objects of interest and taking into account that the synthesized images exhibit a high level of complexity in terms of the depicted sceneries, using edge detectors like Canny or Sobel filter, yields edge maps with proportionally high complexity. In our experiments using both the MLP and U-Net latent edge predictor, this redundancy in detected edges results in notably low values of the recall metric due to the significant amount of false positive pixels.

Prior to comparing the extracted edge maps to the input sketches, it is beneficial to perform an additional preprocessing step. The lines composing the input sketches are handwritten and may not have the desired width for direct comparison to the edge maps extracted from the generated images. To alleviate this issue, we refine the input sketches by using an erosion operation with a structuring element $B$, which, in our case, corresponds to a $3\times 3$ all-ones matrix. The eroded sketch is then used as a reference to calculate the recall metric. Finally, based on the contours of the depicted objects in the input sketch, a bounding box is extracted and used to crop the eroded sketch and the edge map of the generated image. This helps to limit the comparison to the area of interest. 

The recall score was computed using 1400 generated images, with U-Sketch showing superior performance compared to the MLP baseline, as evidenced by the results presented in Table \ref{comparative_analysis_examples}.

\noindent \textbf{User Studies:} \ In order to evaluate the effectiveness of our proposed U-Sketch in the sketch-to-image task, we conducted two comprehensive user studies. These studies were structured to compare the performance of our proposed framework, comprising the U-Net latent edge predictor, against the MLP baseline. The user studies were conducted anonymously, with participants unaware of the methodologies underlying the generation of the presented samples.

In our User Preference Evaluation Study, a cohort of 37 participants was provided with sketches accompanied by textual prompts, alongside outputs generated by both the U-Sketch and MLP frameworks, using the same seed number for fair comparison. The provided examples were randomly selected and participants were not informed about which framework produced each image. Subsequently, participants were tasked with assessing each example, to indicate which of the generated images was (a) more realistic, (b) had better edge fidelity, \ie adherence to the sketch input, and (c) had better overall structural coherence relative to the provided sketch and textual guidance.

Subsequently, 1110 comparative evaluations were collected, reflecting strong user preference for the U-Sketch framework, as noted in Table \ref{comparative_analysis_examples}. Specifically, the samples generated by the U-Sketch framework were adjudged to be more realistic in 60.9\% of evaluations. Moreover, in terms of edge fidelity, our proposed method prevailed over the MLP baseline in 70.5\% of cases. Finally, in terms of overall structural coherence, 70.7\% of evaluations indicated the superiority of our method. In summary, the outcomes of the User Preference Evaluation Study substantiated the superiority of our proposed method over the baseline.

In our User Rating Evaluation Study, a cohort of 31 participants was presented with sketch and textual prompt pairs. Each input was accompanied by a single output, generated either by U-Sketch or MLP framework. Once again, participants had no prior knowledge of the framework used to generate the presented image. Each participant was asked to rate the generated image on a scale of 1 to 5 (bad $\rightarrow$ excellent) with regard to (a) realism, (b) edge fidelity, and (c) overall
structural coherence. Ultimately, 930 evaluations were collected, with an equal distribution between evaluations associated with U-Sketch and MLP.

The findings from the User Rating Evaluation Study are presented in Table \ref{comparative_analysis_examples}, wherein the Mean Opinion Score (MOS) is computed by taking the mean ($\mu$) and standard deviation ($\sigma$) of all responses. Notably, our proposed U-Sketch outperforms the MLP baseline across all assessed criteria. Specifically, our method exhibits an improvement of 20.4\% in realism, 33.8\% in edge fidelity, and 26.1\% in structural coherence relative to the baseline. Furthermore, our approach yields reduced variance in MOS, indicative of heightened consensus among participants regarding the quality of its outputs.
\begin{table}[t]
    \centering
    \caption{Results of quantitative evaluation. }
    \vspace*{-0.6\baselineskip}
    \resizebox{\textwidth}{!}{
    {\setlength{\tabcolsep}{5.8pt}
    \def\arraystretch{1.7 }
    \begin{tabular}{c|C{1.4cm}|c|c|c|c|c|c}
    \hline\hline
     \multirow{2}{*}{\vspace*{-0.5cm}\textbf{Framework}} &  \multirow{2}{*}{\vspace*{-0.23cm} \hspace*{-0.16cm}\textbf{Recall} ($\uparrow$)} & \multicolumn{3}{c|}{\textbf{User Preference Evaluation Study} ($\uparrow$)} & \multicolumn{3}{c}{\textbf{User Rating Evaluation Study} ($\uparrow$)}\\
    \cline{3-8}
    & &  \textit{Realism} & \textit{Edge} \textit{fidelity} &\rule{0pt}{20pt} \shortstack{\textit{Overall structural}\\ \textit{coherence}} & \textit{Realism} ($\pm \sigma$)& \textit{Edge} \textit{fidelity} ($\pm \sigma$)& \shortstack{\textit{Overall structural}\\ \textit{coherence} ($\pm \sigma$)}\\
    \hline\hline
       \textbf{MLP\textsubscript{LEP}}  & \hspace*{-0.16cm} 0.595 & 39.1\% & 29.5\% & 29.3\% & 3.29 $\pm$ 1.04 & 2.89 $\pm$ 1.02 & 3.20 $\pm$ 1\\
       \textbf{U-Sketch}  & \hspace*{-0.16cm} \textbf{0.645} & \textbf{60.9\%} & \textbf{70.5\%} & \textbf{70.7\%} & \textbf{3.97 $\pm$ 1} & \textbf{3.86 $\pm$ 0.95} & \textbf{4.03 $\pm$ 0.93}\\
    \Xhline{0.25ex}
    \end{tabular}}}
    \label{comparative_analysis_examples}
    \vspace*{-1.2\baselineskip}
\end{table}

\section{Conclusion} In this paper we introduce U-Sketch, the first, to the best of our knowledge, diffusion-based sketch-to-image framework, comprising a U-Net architecture for edge map prediction, during inference time. The convolutional nature of U-Net facilitates precise estimation of edge maps for generated images, yielding high-quality samples that closely adhere to the reference sketches. We conducted thorough qualitative evaluations and extensive user studies, solidifying the efficacy and superiority of U-Sketch over existing methods. Our framework achieves promising results with significantly fewer inference steps, demonstrating higher efficiency while preserving image realism.

As future work, there remains a pressing need to develop better evaluation metrics tailored specifically for the sketch-to-image task. Additionally, the impact of initial noise on the quality and structure of generated images warrants further investigation. Understanding and mitigating these factors will undoubte- dly contribute to advancing the field, enhancing the performance and reliability of sketch-to-image frameworks, such as U-Sketch.

\bibliographystyle{splncs04}
\bibliography{main}
\end{document}